\documentclass{vgtc}                          

\ifpdf
  \pdfoutput=1\relax                   
  \usepackage{graphicx}                
  \DeclareGraphicsExtensions{.pdf,.png,.jpg,.jpeg} 
\else
  \usepackage{graphicx}                
  \DeclareGraphicsExtensions{.eps}     
\fi%

\graphicspath{{figures/}{pictures/}{images/}{./}} 

\usepackage{microtype}                 
\PassOptionsToPackage{warn}{textcomp}  
\usepackage{textcomp}                  
\usepackage{mathptmx}                  
\usepackage{amsmath}
\usepackage{amssymb}
\usepackage{times}                     
\usepackage{cite}                      
\usepackage{tabu}                      
\usepackage{booktabs}                  
\usepackage{pifont}
\usepackage[table,xcdraw]{xcolor}
\usepackage{multirow}

\usepackage{indentfirst}

\usepackage{algorithm}
\usepackage{algorithmic}

\onlineid{0}

\vgtccategory{Research}

\vgtcinsertpkg

\title{3D Colored Shape Reconstruction from a Single RGB Image through Diffusion}

\author{Bo Li\thanks{e-mail: libonchu@outlook.com} %
\and Xiaolin Wei
\and Weifeng Chen
\and Bin Liu\thanks{e-mail:nyliubin@nchu.edu.cn}}
\affiliation{\scriptsize School of Mathematics and Information Science \\ Nanchang Hangkong University, Nanchang, China}

\abstract{We propose a novel 3d colored shape reconstruction method from a single RGB image through diffusion model. Diffusion models have shown great development potentials for high-quality 3D shape generation. However, most existing work based on diffusion models only focus on geometric shape generation, they cannot either accomplish 3D reconstruction from a single image, or produce 3D geometric shape with color information. In this work, we propose to reconstruct a 3D colored shape from a single RGB image through a novel conditional diffusion model. The reverse process of the proposed diffusion model is consisted of three modules, shape prediction module, color prediction module and NeRF-like rendering module. In shape prediction module, the reference RGB image is first encoded into a high-level shape feature and then the shape feature is utilized as a condition to predict the reverse geometric noise in diffusion model. Then the color of each 3D point updated in shape prediction module is predicted by color prediction module. Finally, a NeRF-like rendering module is designed to render the colored point cloud predicted by the former two modules to 2D image space to guide the training conditioned only on a reference image. As far as the authors know, the proposed method is the first diffusion model for 3D colored shape reconstruction from a single RGB image. Experimental results demonstrate that the proposed method achieves competitive performance on colored 3D shape reconstruction, and the ablation study validates the positive role of the color prediction module in improving the reconstruction quality of 3D geometric point cloud.

} 

\CCScatlist{
  \CCScatTwelve{Shape reconstruction from image}{Colored point cloud}{Conditional diffusion probability model}{Markov process};
}

\begin{document}


\firstsection{Introduction}

\maketitle

3D shape reconstruction is a basic research topic in computer graphics and computer vision. In the past decades, point cloud has been a popular 3D shape representation due to its simplicity and compactness. 
Although 3D scanning hardware has achieved great development in recent years, it cannot meet the demand of industry. On one hand, it is expensive and time consuming to scan a large amount of real models. On the other hand, in many cases, designers wonder to generate some novel 3D shapes that may not exist in real world. Therefore, 3D shape reconstruction or generation is still an important research issue in computer 
vision.

Generative neural networks, such as generative adversarial networks (GANs)~\cite{goodfellow2020generative}, have gained competitive performance on image synthesis~\cite{zhu2017unpaired,emami2020spa,gu2020image}, and have been extended to 3D shape generation~\cite{chan2022efficient,wang2017shape,shu20193d,li20193d}. Despite of the powerful generation potential, GANs have some inherent limitations, e.g., the training procedure could be unstable due to the adversarial training. In recent years, a new class of generative model, referred to diffusion probability model~(DPM)\cite{sohl2015deep,ho2020denoising}, has been popular and beated GANs in numerous fields, such as image generation~\cite{dhariwal2021diffusion,ho2022cascaded,vahdat2021score,meng2021sdedit}, image-to-image translation~\cite{saharia2022palette,choi2021ilvr,gu2022vector}, etc.. As the training of DPM is a simple max-likelihood estimation, it is much more robust than the training of GANs. 

DPM has been applied for 3D shape generation~\cite{cai2020learning,luo2021diffusion,zhou20213d}. In~\cite{cai2020learning}, 3D point cloud is generated through a score-based model~(ShapeGF) which can be seen as a variant of DPM under some mild conditions. The method strikes to first find the optimal score function, i.e., the gradient of the logarithm probability distribution of the training data, and then according to Langevin sampling theory, one can generate a high-fidelity sample in the data distribution from a random noise. \cite{luo2021diffusion} proposed a novel latent shape conditioned probabilistic generative model~(DPC) for point clouds by combining variational autoencoder and DPM. The method trained a variational autoencoder to extract the latent shape distribution and then a random sampled shape vector will be used as a condition to guide the reverse training of DPM. The variational autoencoder and DPM are trained simultaneously. \cite{zhou20213d} proposed a point-voxel Diffusion model~(PVD) for point cloud generation. The method can handle unconditional shape generation and conditional shape completion. 

\begin{figure}
\centering
\includegraphics[width=1\linewidth]{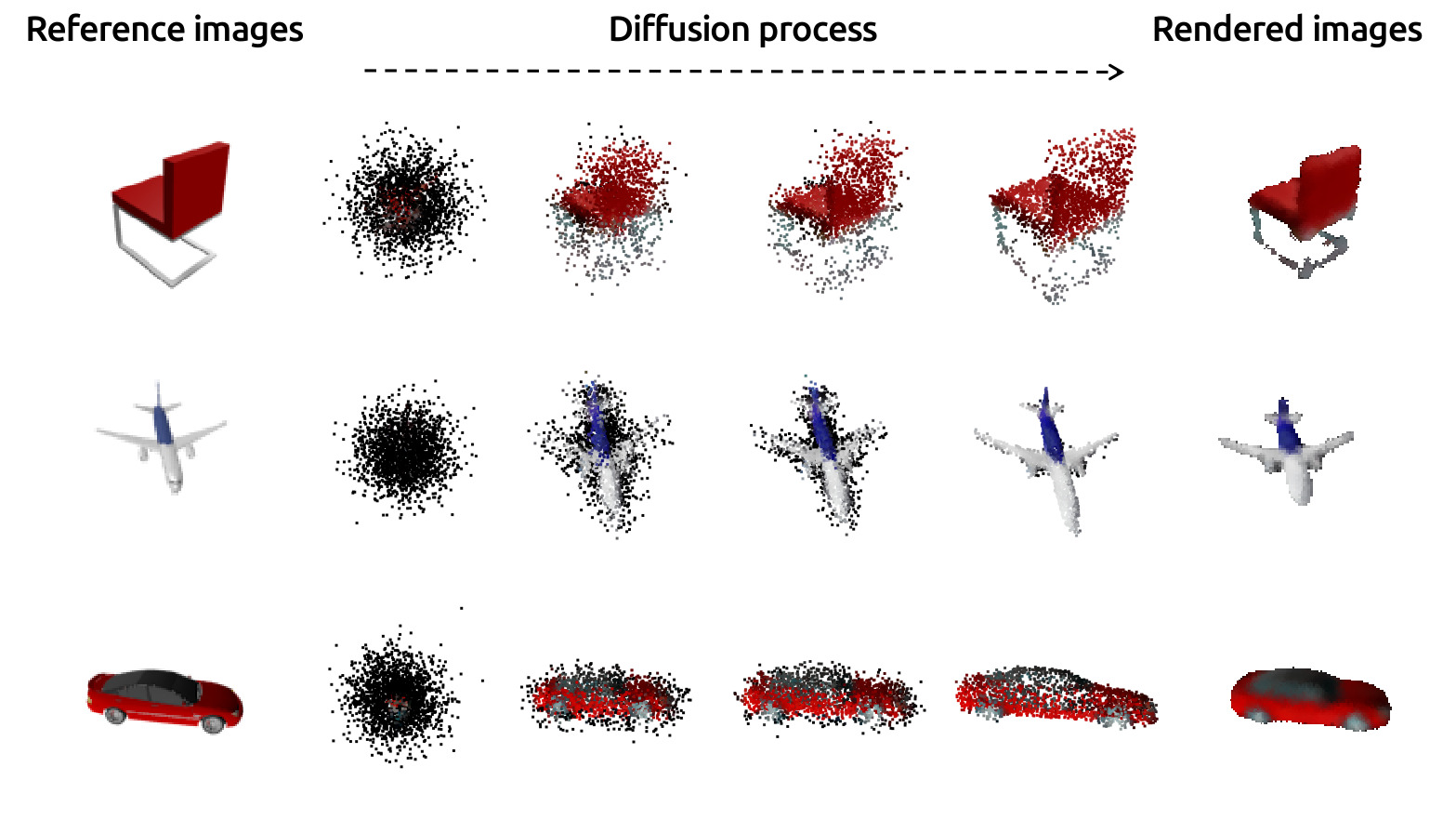}
\caption{Point clouds reconstruction from a single RGB image with the proposed diffuse method. Given a reference image~(left column), the proposed diffusion model will gradually reconstruct a colored point cloud corresponding to the reference image. Finally, the rendered image of the reconstructed point cloud is shown in the last column. }
\label{first}
\end{figure}

\begin{figure*}
\centering
\includegraphics[width=1\linewidth]{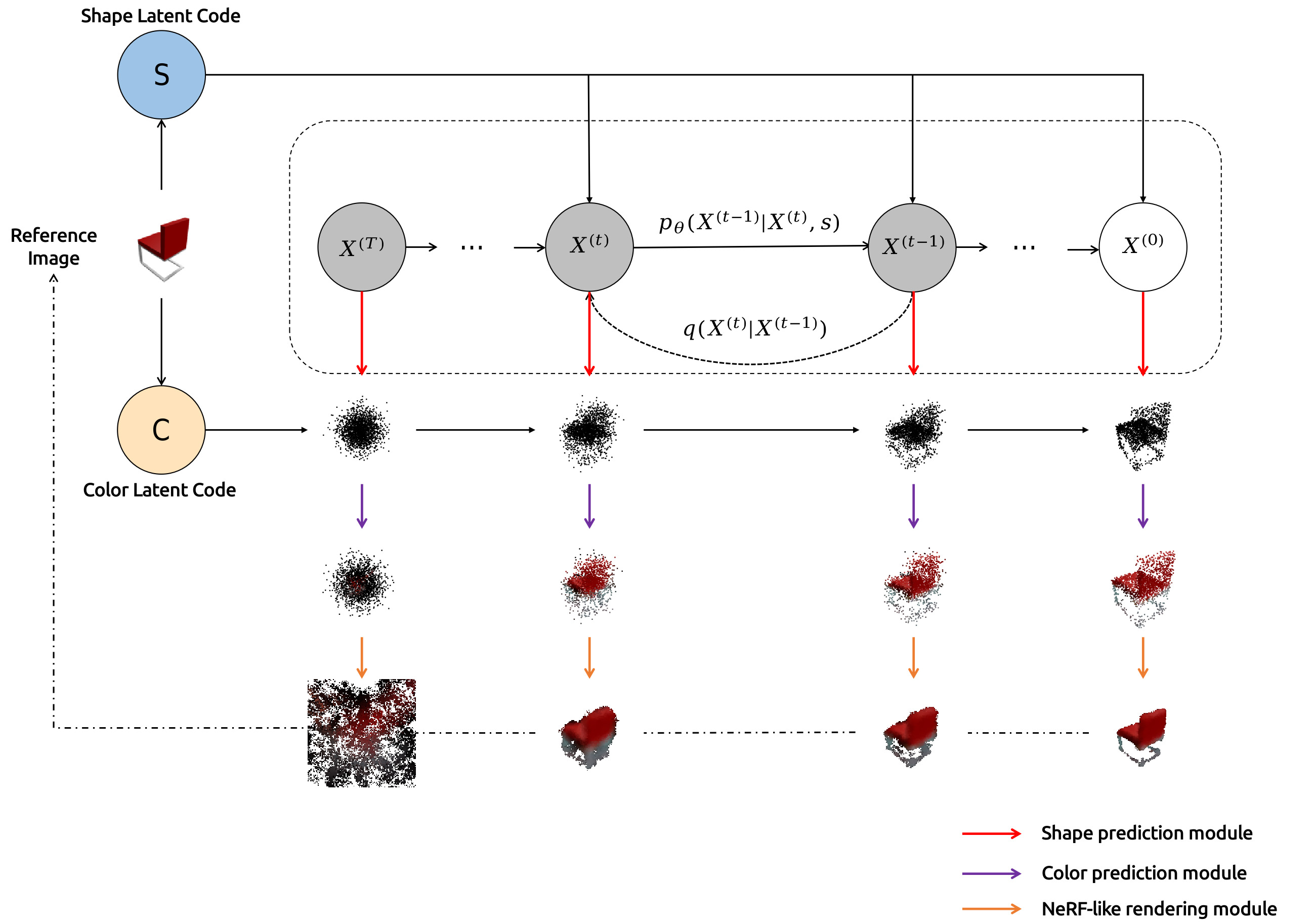}
\caption{Our overall framework is composed of three modules: shape prediction module, color prediction module and NeRF-like rendering module. The reference image is first encoded into a shape latent code $\mathbf{S}$ and a color code $\mathbf{C}$, and then the shape feature is fed to the shape prediction module to reconstruct the geometric shape of point cloud. Then the color feature is utilized to predict the color of each 3D point. Finally the NeRF-like rendering module is proposed to render the colored point clouds to 2D RGB image space.}
\label{model}
\end{figure*}

While remarkable progress has been made, the existing DPM based methods~\cite{cai2020learning,luo2021diffusion,zhou20213d} only focus on geometric shape generation, they cannot either accomplish 3D reconstruction from a single image, or produce 3D geometric shape with color information. 
In this work, we propose to reconstruct a 3D colored shape from a single RGB image through a novel conditional diffusion model. Similar to DPM, the forward process of the proposed diffusion model converts the initial distribution of the point cloud to a simple Gaussian distribution via a Markov process. In order to accomplish colored point cloud reconstruction from a single RGB image, the reverse process of the proposed diffusion model is consisted of three modules, shape prediction module, color prediction module and NeRF-like rendering module. In shape prediction module, the RGB image is first encoded into a high-level shape feature and then the shape feature is utilized as a condition to predict the reverse geometric noise in diffusion model. Then the color prediction module is designed to predict the RGB color of each 3D point of the updated point cloud. As there is only a single RGB image as reference for point cloud reconstruction, and no ground truth colored 3D point cloud available, a NeRF-like rendering module is proposed to render the colored point cloud to 2D image, and the loss between the rendered image and the ground truth reference image will be used to guide the  training process. Some examples are shown in Fig.~\ref{first} to illustrate the reconstruction process and results of the proposed method. Given a reference image as shown in the first column of Fig.~\ref{first}, the proposed diffusion model will gradually reconstruct a colored point cloud from a random noise sampled from Gaussian distribution to a meaningful colored shape corresponding to the reference image. Finally, in order to illustrate the reconstruction quality, the rendered image of the reconstructed point cloud is shown in the last column. 
As far as the authors know, the proposed method is the first diffusion model for 3D colored shape reconstruction from a single RGB image.

The main contributions of the proposed method include,
\begin{itemize}
\item {We proposed a novel diffusion model for colored point cloud reconstruction conditioned on a single RGB image. As far as the authors know, it is the first diffusion model designed for 3D colored shape generation from a 2D image.}

\item{The proposed model cannot only reconstruct 3D geometric information of point cloud, but also predict the color of each point with the guidance of the reference RGB image.}

\item{Extensive experiments show that our model achieves
competitive performance in colored point cloud reconstruction from a single RGB image.}

\end{itemize}

\section{Related Work}

\textbf{Diffusion Probability Model} learns the probability distribution of training data through two dual stochastic process. The forward diffusion process converts original data distribution to a simple normal distribution by gradually adding Gaussian noise to the ground truth shape. In fact, it is a Markov process with Gaussian transition probability. The theory of stochastic process guarantees the tractable transformation. The forward process is free of training, it provides a reference ground truth trajectory for the reverse process.
The reverse process is trained to learn the optimal denoising strategy to constrain the reverse trajectory coincide with the forward process. Intuitively, the reverse diffusion process learns a stochastic path from Gaussian distribution to the data distribution. The diffusion probability model can be described mathematically as follows.

The forward process from data $x_0 \sim q_{data}(x_0)$ to the latent variable $x_T$ can be formulated as a fixed Markov chain:
\begin{align}
    q(x_1, ..., x_T | x_0) = \prod_{t=1}^T q(x_t | x_{t-1}) \label{DDPMF}
\end{align}
where $q(x_t | x_{t-1}) = \mathcal{N}(x_t; \sqrt{1-\beta_t}x_{t-1}, \beta_t I)$, $\beta_t$ is a small positive constant, $T$ is the largest number of transition. The forward process gradually perturbs original data $x_0$ to a latent variable with an isotropic Gaussian distribution $\mathcal{N}(0, I)$. According to the Kolmogorov equation, the sampling distribution of $x_t$ can be directly derived from $x_0$:
\begin{align}
    q(x_t | x_0) = \mathcal{N}(x_t; \sqrt{\Bar{\alpha_t}}x_0, (1-\Bar{\alpha_t})I)
\end{align}
where $\alpha_t = 1-\beta_t$ and $\Bar{\alpha_t}=\prod_{s=1}^t \alpha_s$

The reverse process strives to predict the original data $x_0$ from the latent variable $x_T \sim \mathcal{N}(0, I)$ through a dual Markov chain:
\begin{align}
    p_{\theta}(x_0, ..., x_{T-1} | x_T) =  \prod_{t=1}^T p_{\theta}(x_{t-1} | x_t) \label{DDPMR}
\end{align}
The training objective of DDPM is to optimize ELBO. 
\begin{align}
    & E_q\big[D_{KL}(q(x_T | x_0) || p(x_T)) \notag \\
    & + \sum_{t>1}D_{KL}(q(x_{t-1} | x_t, x_0) || p_{\theta}(x_{t-1} | x_t)) \notag \\
    & - log p_{\theta}(x_0 | x_1)\big]
\end{align}
Combining Eq.(\ref{DDPMF}) and Eq.(\ref{DDPMR}), the objective can be simplified as optimizing:
\begin{align}
    \mathbb{E}_{x_0, \epsilon}||\epsilon-\epsilon_{\theta}(x_t, t)||^2_2
    \label{loss_dif}
\end{align}
where $\epsilon$ is the Gaussian noise in $x_t$, $\epsilon_{\theta}$ is the model trained to estimate $\epsilon$.

\textbf{3D Shape Reconstruction} 
Traditional methods mainly contain two kinds of techniques in 3D shape reconstruction. One kind of techniques utilize depth information of objects with the help of scanning device, such as KinectFusion~\cite{izadi2011kinectfusion} and BundleFusion~\cite{dai2017bundlefusion}. The other uses the structure consistency information which can be built by feature point matching strategy among multi-view images, such as MVS~\cite{furukawa2015multi} and COLMAP~\cite{schonberger2016structure}. However, the operation process of these methods is too complex and many assumptions are required, such as the cover among views. Therefore, researchers focus on seeking a simple and straightforward technique, namely reconstructing 3D shapes from one single RGB image directly. 
 

With the development of deep learning, researchers catch sight of new dawn through the data-driven techniques. Sun et al.~\cite{sun2019ssl} propose a self-supervised network to generate 3D point clouds from a single RGB image, unlike the existing work which requires multiple views of the same object to recover the full 3D geometry. Fan et al.~\cite{fan2017point} introduce a conditional shape sampler, which can predict multiple plausible 3D point clouds from an input image. Mandikal et al.~\cite{mandikal20183d} first learn a latent embedding matching among 3D shapes then mapping the 2D image to the corresponding embedding to achieve 3d shape reconstruction form single view image. 
Further, they bring up a pyramid structure to obtain a dense point cloud~\cite{mandikal2019dense}. 
Lin et al.~\cite{lin2018learning} use 2D convolutional operations to predict the 3D structure from multiple viewpoints and jointly apply geometric reasoning with 2D projection optimization to obtain result efficiently. In order to get colored point cloud, Hu et al.~\cite{hu2021learning} introduce a two-stage approach, which first infers an object coordinate map from the input RGB image, and then obtain the final point cloud by reprojection and completion.


Although previous works have achieved success in different aspects, their accuracy is still not satisfactory. Recently, diffusion based models by denoising step by step show remarkable performance in obtaining better effect no matter on what metrics. 
Cai et al.~\cite{cai2020learning} perform stochastic gradient ascent on an unnormalized probability density, then move sampled points toward the high-likelihood regions to get high quality point cloud. Luo et al.~\cite{luo2021diffusion} derive a variational bound in closed form for training reverse diffusion process for point clouds conditioned on a shape latent space. Zhou et al.~\cite{zhou20213d} marries denoising diffusion models with the hybrid, point-voxel representation of 3D shapes for unconditional shape generation and conditional, multi-modal shape completion. However, the existing DPM based methods~\cite{cai2020learning,luo2021diffusion,zhou20213d} only focus on geometric shape generation, they cannot either accomplish 3D reconstruction from a single image, or produce 3D geometric shape with color information. 
In this work, we propose to reconstruct a 3D colored shape from a single RGB image through a novel conditioned diffusion model. As far as the authors know, it is the first diffusion model designed for 3D colored shape generation from a 2D image.


\section{Methods}
In this section, we introduce the main idea of the proposed 3D colored point cloud reconstruction from a single RGB image. The framework of the proposed method is shown in Fig.\ref{model}. 

Compared with the existing diffusion based point cloud generation methods, the proposed method can not only accomplish point cloud reconstruction from a single image, but also can predict the RGB color of each 3D point. To achieve the above objectives, the reverse process of the proposed diffusion model is consisted of three modules, shape prediction module, color prediction module and NeRF-like rendering module. Shape prediction module is designed to reconstruct 3D geometric structure of point cloud from a 2D image, while color prediction module is to predict the color of each 3D point updated in the former shape prediction module. As there is only a single RGB image as reference for point cloud reconstruction, and no ground truth colored 3D point cloud available, a NeRF-like rendering module is proposed to render the colored point cloud to 2D image, and the loss between the rendered image and the ground truth reference image will be used to guide the  training process. The total loss function is composed of geometric reconstruction error and color difference between reference RGB image and the rendered result by NeRF.  


\subsection{Formulation}
Given a reference RGB image $\mathbf{I}$, the proposed method aims at reconstructing the corresponding 3D point cloud $\mathbf{X} \in \mathbf{R}^{N\times 3}$ and the color of each 3D point $\mathbf{\mathcal{C}}\in \mathbf{R}^{N\times 3}$, where $N$ is the number of points of point cloud object. It is noticed that in the training stage, the reference RGB image $\mathbf{I}$ and its corresponding 3D point cloud $\mathbf{X}$ will be given in pair as training data.

In the forward diffusion process, the original point cloud data $\mathbf{X}^{(0)}$ will be transited to a pure  noise by gradually adding Gaussian noise, 
\begin{equation}
    \mathbf{X}^{(t)} = \sqrt{1-\beta_t}  \mathbf{X}^{(t-1)} + \sqrt{\beta_t} n
    \label{forward}
\end{equation}
where $\beta_t$ is a small positive constant, $t$ denotes the time step, and $n$ is a random noise sampled from standard normal distribution. The forward diffusion can also be regarded as a Markov chain with transition probability 
$q(x_t | x_{t-1}) =\mathcal{N} (x_t; \sqrt{1-\beta_t}x_{t-1}, \beta_t I)$. With the above transition probability, the Markov process will converge to a stationary distribution after infinite number of transitions, i.e., Gaussian distribution. In such a way, a piece of stochastic path composed of noisy point clouds $\{\mathbf{X}^{(i)}\}, i=0, 1, 2\cdots, T$ will be generated. The largest transition number $T$ is set to 200 in this paper. At time $T$, the point cloud $\mathbf{X}^{(T)}$ is almost a random noise.

In the reverse diffusion process, given a random noise $\mathbf{X}^{(T)}$ sampled from Gaussian distribution, the proposed method aims at predicting both the geometric structure and color information of the 3D point cloud corresponding to the reference RGB image $\mathbf{I}$. As shown in Fig.~\ref{model}, the reverse process of the proposed diffusion model is consisted of three modules: shape prediction module, color prediction module and NeRF-like rendering module. The reference image is first encoded into a shape latent code $\mathbf{S}$ and a color code $\mathbf{C}$, and then the shape feature is fed to the shape prediction module to reconstruct the geometric shape of point cloud. Then the color feature $\mathbf{C}$ is utilized to predict the color of each 3D point updated in shape prediction module. Finally a NeRF-like rendering module is proposed to render the colored point clouds to 2D RGB image space to match the ground truth reference image $\mathbf{I}$.


\begin{figure}
\centering
\includegraphics[width=1\linewidth]{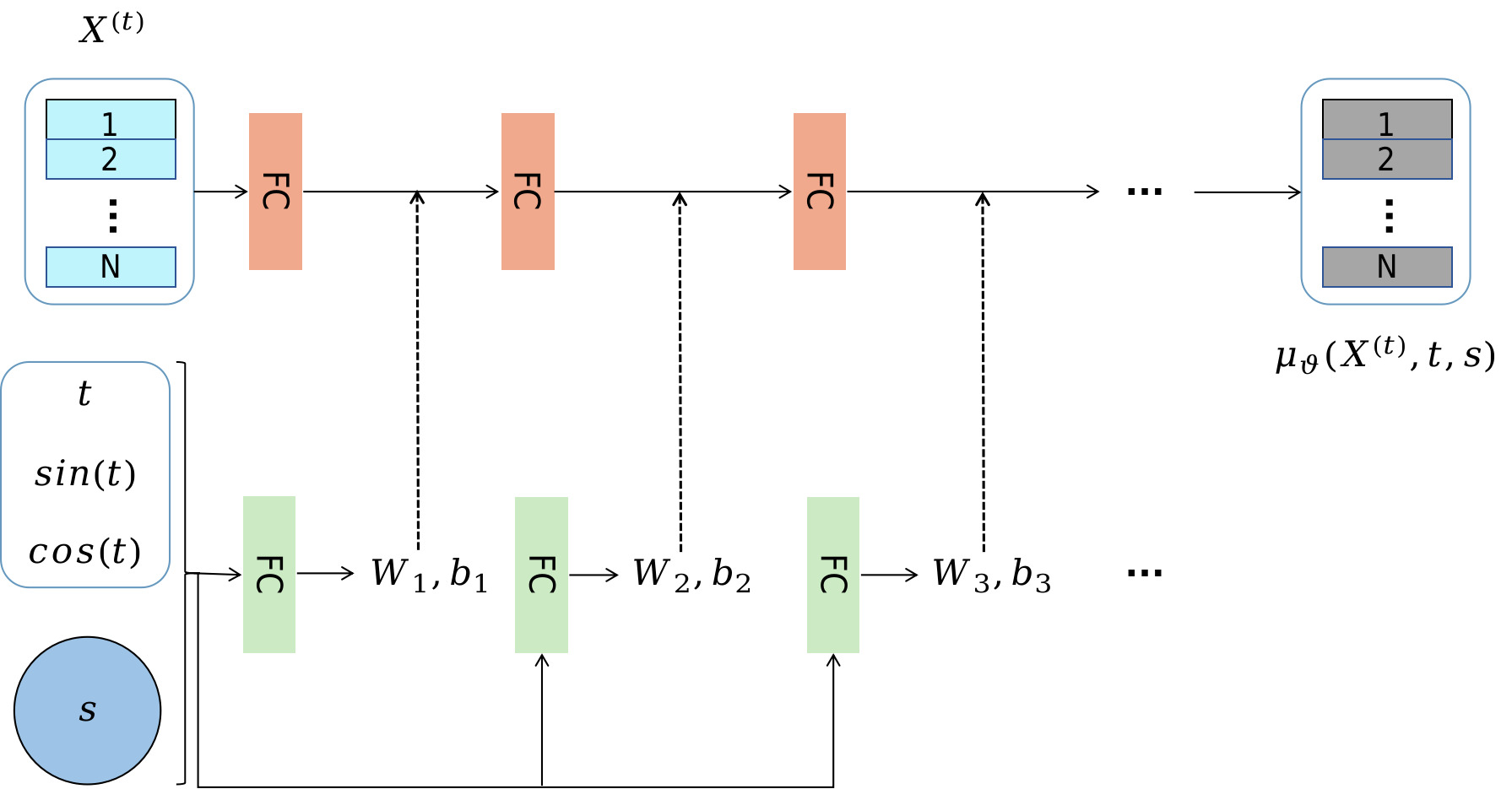}
\caption{The details of shape prediction module. In the figure we have omitted the activation function.}
\label{pred}
\end{figure}

\textbf{shape prediction module} is to reconstruct the geometric shape of point cloud corresponding to the object in the reference image.
In this paper, we model shape prediction as a diffusion process conditioned on shape feature $\mathbf{S}$ encoded from the reference image. 

Given a random noise from Gaussian distribution, shape prediction module aims at determining a piece of path from the noise to the ground truth point cloud corresponding to the object in the reference image. Mathematically, it attempts to explore the optimal transition probability in the reverse Markov diffusion process. 
Specifically, we formulate the shape module of the inverse diffusion process as:
\begin{equation}
p_{\mathbf{\theta}}\left(\mathbf{X}^{(0:T)} \mid \mathbf{S}\right)=p\left(\mathbf{X}^{(T)}\right) \prod_{t=1}^T p_{\mathbf{\theta}}\left(\mathbf{X}^{(t-1)} \mid \mathbf{X}^{(t)}, \mathbf{S}\right)
\end{equation}
\begin{equation}\label{shapepred}
p_{\mathbf{\theta}}\left(\mathbf{X}^{(t-1)} \mid \mathbf{X}^{(t)}, \mathbf{S}\right)=\mathcal{N}\left(\mathbf{X}^{(t-1)} \mid \mathbf{\mu}_{\mathbf{\theta}}\left(\mathbf{X}^{(t)}, t, \mathbf{S}\right), \beta_t \boldsymbol{I}\right)
\end{equation}
where $\mathbf{\mu}_{\mathbf{\theta}}$ is a neural network parameterized by $\mathbf{\theta}$, and it is designed to predict the mean value of the noise. $\mathbf{S}$ is the shape feature extracted from the reference image. The specific architecture of the network is shown in Fig.\ref{pred}. 
In order to guide the diffusion direction of the point cloud, we propose to modulate the noise prediction function by the guidance of shape feature $\mathbf{S}$. Specifically, each layer of noise prediction function is modulated by gate-bias modulation with different parameters. In each modulation stage, we concatenate the shape feature $\mathbf{S}$ with the time variable $t$ and fourier embedding of $t$ as condition, then we feed the condition into two different linear layers to generate gate and bias, respectively. Note that the gate tensor is further squeezed through a sigmoid non-linearity, and perform per-dimensional scaling operation on the input feature. Then the bias tensor translates the scaled feature along a particular edit direction through per-dimensional addition operations. Finally, the output is passed through a non-linearity activation before passing to the next modulation layer.

Given a reference image $\mathbf{I}$ and an initial point cloud $\mathbf{X^{(T)}}$ sampled from a Gaussian distribution, we can reconstruct the geometric shape of the point cloud corresponding to the object in the reference image  $\mathbf{I}$ through shape prediction module.

\textbf{color prediction module} is to predict the color of each point updated in shape prediction module. As there is no ground truth color supervision at intermediate time of the diffusion process, we propose to predict the color of the final diffusion result in the reverse process with the guidance of the ground truth reference image. Benefit from the theory of multi-step Kolmogorov equation, the diffusion state at time $t$ can be tracked by the following formulation from initial time.
\begin{equation}
\mathbf{X}^{(t)}=\sqrt{\bar{\alpha}_t} \mathbf{X}^{(0)}+\sqrt{1-\bar{\alpha}_t}n
\end{equation}
where $\alpha_t:=1-\beta_t \text { and } \bar{\alpha}_t:=\prod_{s=1}^t \alpha_s$. In this way, from the predicted shape $\bar{\mathbf{X}}^{(t)}$ at time $t$, we can induce the approximated final state by $\bar{\mathbf{X}}^{(0)}=\frac{1}{\sqrt{\bar{\alpha}_t}} \bar{\mathbf{X}}^{(t)} - \frac{\sqrt{1-\bar{\alpha}_t}}{\sqrt{\bar{\alpha}_t}}n$.

Given the diffusion result $\bar{\mathbf{X}}^{(t)}$ at time $t$, a color prediction neural network $\phi_{\theta}$ is designed to predict the color $\mathcal{C}$ of each point conditioned on the latent color feature $\mathbf{C}$ extracted from the reference image,

\begin{equation}\label{colorpred}
    \mathcal{C}_{t,N} = \Gamma_{\varphi}(\bar{X}^{(0)}_{t}, \mathbf{C})
\end{equation}
where $\varphi$ is the parameters of color prediction neural network, and $N$ is the number of point cloud. The specific architecture of the network is similar to that of shape prediction module as shown in Fig.\ref{pred}.

\textbf{NeRF-like rendering module.}
As there is only a 2D reference image $\mathbf{I}$ as color supervision, and no colored 3D point cloud available in the training process, we propose to first render the predicted colored point cloud to 2D image space, and constrain the training by the loss between the rendered image and the ground truth reference RGB image.
In this paper, a NeRF-like rendering strategy is proposed. Specifically, the radiance of a pixel can be measured by passing a ray across the pixel, sampling $M$ shading points $\{\mathbf{p_i}|i=1,2,...,M\}$ along the ray in space, and accumulating them using the volume density, defined as follows:
\begin{equation}
    \begin{aligned}
\mathbf{I}^{\prime} &=\sum_M \tau_i\left(1-\exp \left(-\sigma_i \Delta_i\right)\right) \mathcal{C}_i \\
\tau_i &=\exp \left(-\sum_{t=1}^{i-1} \sigma_t \Delta_t\right)
\label{nerf}
\end{aligned}
\end{equation}
where $\sigma_i$ and $\mathcal{C}_i$ represent volume density and radiance for each shading point, $\tau$ indicates volume transmittance, and $\Delta_i$ is the distance between adjacent shading samples. 
Once the volume density and radiance at each sampled shading point are known, the neural radiation field can render an image $\mathbf{I}^{\prime}$ of the corresponding viewpoint through formulation~(\ref{nerf}). 

At each time step of the reverse diffusion process, the density and radiance of each sampling point can be derived from the updated point cloud.
Given any spatial location $\mathbf{p_i}$, we are querying the $K$ neighboring points in $\mathbf{\bar{X}^{(0)}_{t}}$ around it, denoted as $\{\mathbf{x}_{i,j}^{(t)}|j=1,2,...,K\}$. Then the volume density $\sigma_i$ can be computed by
\begin{equation}
    \sigma_i=1 /\Vert p_i-\frac{1}{K} \sum_{j=1}^K \mathbf{x}_{i, j}^{(t)}\Vert_2
\end{equation}
where $||\cdot||_2$ denotes the $\ell_2$ norm. Similarly, the radiance $\mathcal{C}_i$ can be formulated as:
\begin{equation}
    \mathcal{C}_i=\sum_{j=1}^K \frac{\exp(-\Vert p_i-x_{i,j}^{(t)}\Vert_2)}{\sum_{j=1}^K \exp(\Vert p_i-x_{i, j}^{(t)}\Vert_2)} \mathcal{C}_{t,j}
\end{equation}
where $\mathcal{C}_{t,j}$ is the color of the $j$th point at step $t$ predicted by the color prediction module. Finally, according to Eq.\ref{nerf}, we can get a rendered image $\mathbf{I}_t^{\prime}$ at this viewpoint. In addition, we perform a masking operation on the sampled points in space to speed up the training by discarding those far away from the current point.

It is noticed that there is no trainable parameters in the NeRF-like rendering module.


\begin{figure*}
\centering
\includegraphics[width=1\linewidth]{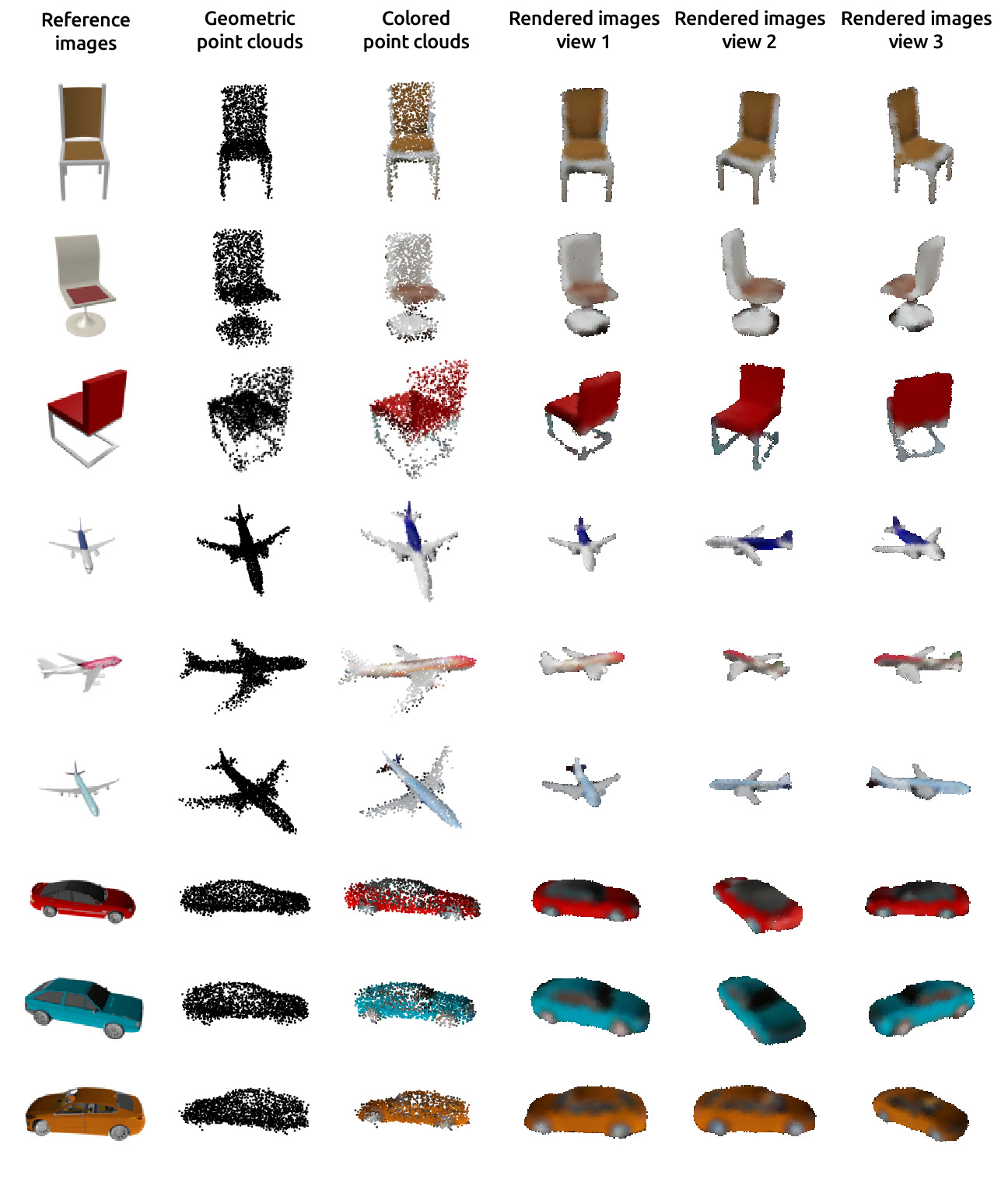}
\caption{Demonstration of colored point cloud reconstruction result from  reference images. The first column is the reference image, the predicted geometric point cloud without and with color are respectively shown in the second and third column. The rendered images from different viewpoints based on the reconstructed colored point cloud are shown in the last three columns.}
\label{result}
\end{figure*}

\begin{figure}
\centering
\includegraphics[width=1\linewidth]{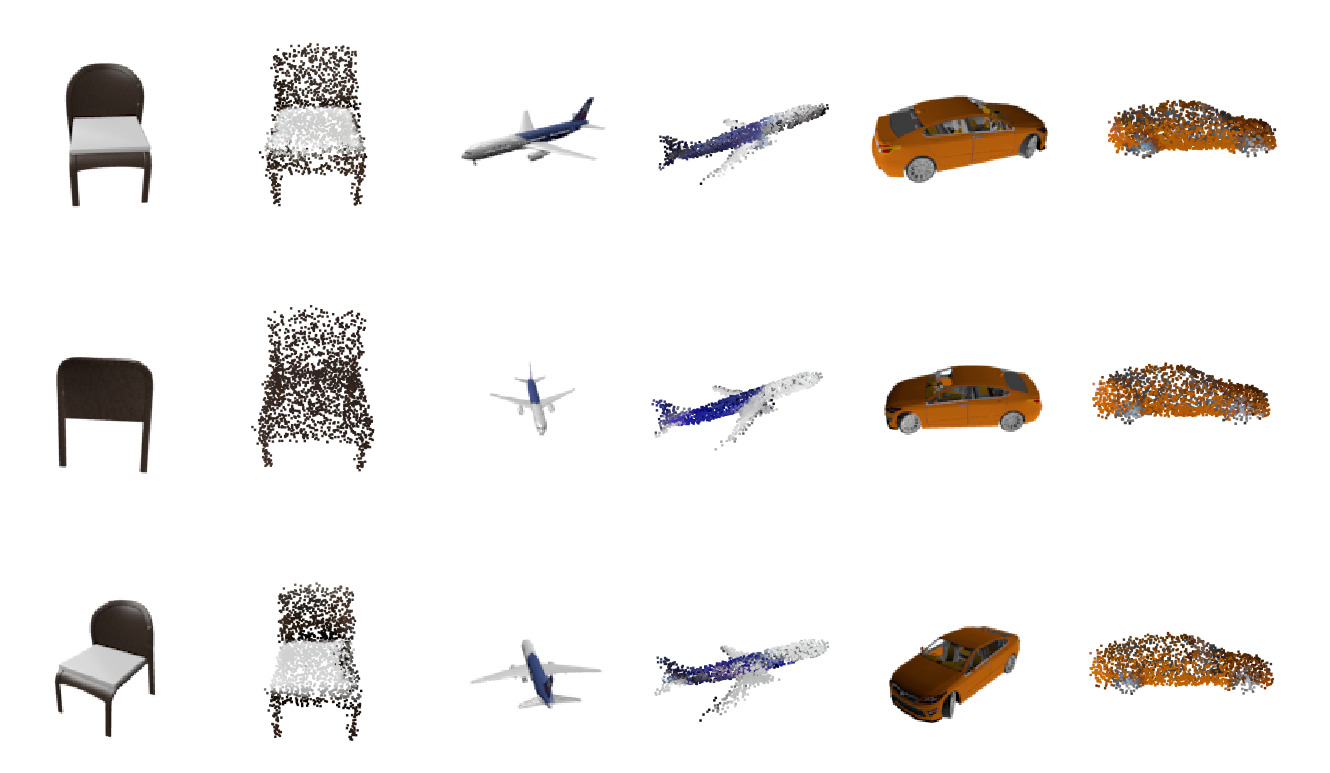}
\caption{Point clouds reconstructed from different viewpoints of the same object.}
\label{cons}
\end{figure}

\subsection{Training Objective}
The training objective of the proposed method is composed of the following four loss functions.

\ding{172} The first loss function is the variational lower bound $\mathbf{\mathcal{L}}_{geo}$ of the reverse diffusion process in shape prediction module. 
\begin{eqnarray*}
     \mathcal{L}_{geo} &=& \mathbf{E}_{q}[\mathbf{D}_{KL}(q(X^{(0)}|X^{(T)})\|p_{\theta}(X^{(T)})) - \log p_{\theta}(X^{(0)}|X^{(1)}) \\ &+& \sum_{t=2}^{T} \mathbf{D}_{KL}(q(X^{(t-1)}|X^{(t)}, X^{(0)})
    \| p_{\theta}(X^{(t-1)}|X^{(t)}, \mathbf{S}))]
\end{eqnarray*}
where $p_{\theta}$ is the shape prediction network in the reverse diffusion process, and $q(X^{(t-1)}|X^{(t)})$ is the transition probability in the forward diffuse, $\mathbf{S}$ is the latent shape feature extracted from the reference image. 

According to the denoising diffusion probability model~\cite{ho2020denoising}, the variational lower bound $\mathbf{\mathcal{L}}_{geo}$ can be simplified as 
\begin{equation*}
\mathbf{\mathcal{L}}_{geo}=\mathbb{E}^{X^{(0)}, \epsilon}||\epsilon-\mu_{\theta}(X^{(t)}, t, \mathbf{S})||^2_2,
\end{equation*}
where $\epsilon$ is the Gaussian noise in $X^{(t)}$. 

\ding{173} The second one is the geometric reconstruction loss. In this papaer, we utilize the Chamfer Loss~\cite{fan2017point} to measure the geometric consistency between the predicted 3D point cloud $\bar{\mathbf{X}}^{(0)}_{t}$ from time $t$ and the real point cloud $\mathbf{X}^{(0)}$, i.e., $\mathbf{\mathcal{L}}_{cham}(\mathbf{\bar{\mathbf{X}}^{(0)}_{t}}, \mathbf{X}^{(0)})$.

\ding{174} The third loss is the color reconstruction measurement with respect to the ground truth reference image. The loss function is defined as the Euclidean distance between the rendered image $\mathbf{I}_t^{\prime}$ by NeRF-like rendering module and the ground truth image $\mathbf{I}$, i.e.,  $\mathbf{\mathcal{L}}_{rgb}=\|\mathbf{I}_t^{\prime}- \mathbf{I}\|_{2}^{2}$.


The final total loss is defined as the sum of the above losses, 
\begin{equation}\label{loss_global}
    \mathcal{L} = \mathcal{L}_{geo} + \mathcal{L}_{cham} + \mathcal{L}_{rgb}
\end{equation}

\begin{algorithm}[htb]\label{beta}
\caption{3D colored point cloud reconstruction from a single RGB image.}
\label{alg:Framwork}
\begin{algorithmic}[1]
\REQUIRE ~~\\
{\bf Training data:} point cloud $\mathbf{X}$, the corresponding reference RGB image $\mathbf{I}$\\
{\bf Initialize}: $\mathbf{X}^{(0)} =  \mathbf{X}, \beta_{0} = 1e-4, \beta_{T} = 0.05, T = 200, \mathbf{\theta}, \varphi$.

{\bf Forward diffusion}: $\mathbf{X}^{(t)} = \sqrt{1-\beta_t}  \mathbf{X}^{(t-1)} + \sqrt{\beta_t} n$ ~ (Eq.~\ref{forward})
{\bf Reverse diffusion}:
\REPEAT
\STATE {\bf Image encoder}\\
$(\mathbf{S}, \mathbf{C}) = encoder (\mathbf{I})$
\STATE {\bf Shape prediction module}\\
$p_{\mathbf{\theta}}\left(\mathbf{X}^{(t-1)} \mid \mathbf{X}^{(t)}, \mathbf{S}\right)=\mathcal{N}\left(\mathbf{X}^{(t-1)} \mid \mathbf{\mu}_{\mathbf{\theta}}\left(\mathbf{X}^{(t)}, t, \mathbf{S}\right), \beta_t \boldsymbol{I}\right)$~(Eq.~\ref{shapepred})

\STATE  {\bf Color prediction module}\\ $\mathcal{C}_{t,N} = \Gamma_{\varphi}(\bar{X}^{(0)}_{t}, \mathbf{C})$~(Eq.~\ref{colorpred})\\
\STATE  {\bf NeRF-like rendering module}\\  
$\mathbf{I}^{\prime} =\sum_M \tau_i\left(1-\exp \left(-\sigma_i \Delta_i\right)\right) \mathcal{C}_i$~(Eq.~\ref{nerf})

\STATE {\bf Loss objective:}\\
$ \mathcal{L} = \mathcal{L}_{geo} + \mathcal{L}_{cham} + \mathcal{L}_{rgb}$~(Eq.~\ref{loss_global})
\UNTIL convergence
\ENSURE  Given a reference image $\mathbf{I}$, and a random noise $\mathbf{X}^{(T)}$, output the colored point cloud $(X^{(0)}, \mathcal{C}) \in \mathbf{R}^6$ 
\end{algorithmic}
\end{algorithm}

\section{Experiments}
In this section, numerical experiments are designed to evaluate the
performance of the proposed method by both visual inspection and quantitative evaluation.

\subsection{Experimental Setup}

\textbf{Datasets.}
ShapeNet~\cite{chang2015shapenet} dataset processed by R2N2~\cite{choy20163d} method is employed in our experiments. All of the methods are trained and tested on three categories, including chair, airplane and car. 2048 points are sampled for each point cloud shape, and are normalized to vectors with zero mean and unit variance.

\textbf{Evaluation Metrics.}
Following prior works~\cite{achlioptas2018learning}, we use the Chamfer Distance (CD) and the Earth Mover’s Distance (EMD) to evaluate the reconstruction quality of the point clouds.

\textbf{Baselines.}
In this paper, we employ the related 3D point clouds generation methods based on diffusion probability model as the baselines, including DPC~\cite{luo2021diffusion}, ShapeGF~\cite{cai2020learning} and PVD~\cite{zhou20213d}.

\begin{figure}
\centering
\includegraphics[width=1\linewidth]{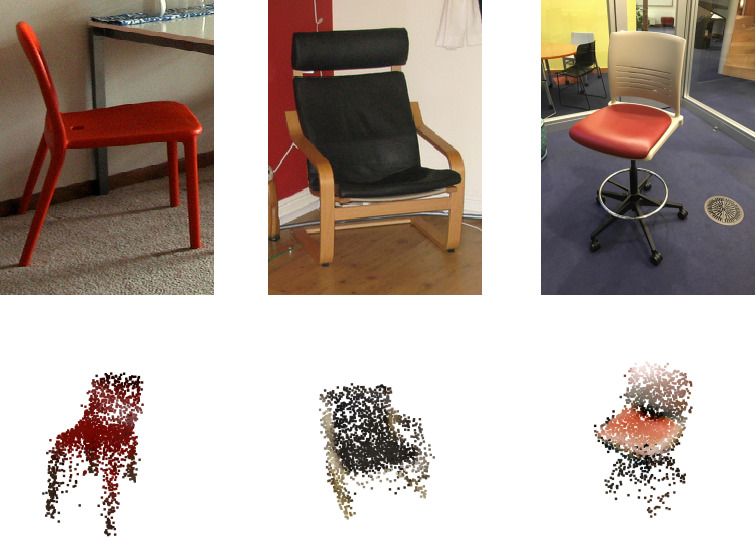}
\caption{Colored point clouds reconstructed from wild images in Pix3D\cite{sun2018pix3d}.}
\label{pix}
\end{figure}

\textbf{Implementation Details}
We train the network by using the Adam optimization algorithm running on a PC with an Intel(R) Core(Tm) i9-9900K CPU @ 3.2 GHz, 32GB of RAM, and a GeForce GTX 3090 GPU.

\subsection{Results.}
\textbf{Colored point cloud reconstruction.} 
First, we evaluate the performance of colored point clouds reconstruction from a single reference RGB image. Experimental results are shown in Fig.~\ref{result}. Given the reference RGB image~(1st column), the reconstructed geometric point cloud and colored point cloud are shown in the 2nd and 3rd column. Finally, in order to evaluate the performance of reconstructed results comprehensively, we demonstrate the rendered images from different views in the last three columns. From Fig.~\ref{result}, we can see that both the geometric and color of the reconstructed point clouds are consistency to the reference RGB image. The third column demonstrate the rendered images with the same view as the reference images, and the similarity between the first column and the third column validates the performance of colored 3D point cloud reconstruction. The rendered images from different view as shown in the last two columns indicate that the proposed method can predict the global point cloud shape well from a single RGB image.

\begin{figure}
\centering
\includegraphics[width=1\linewidth]{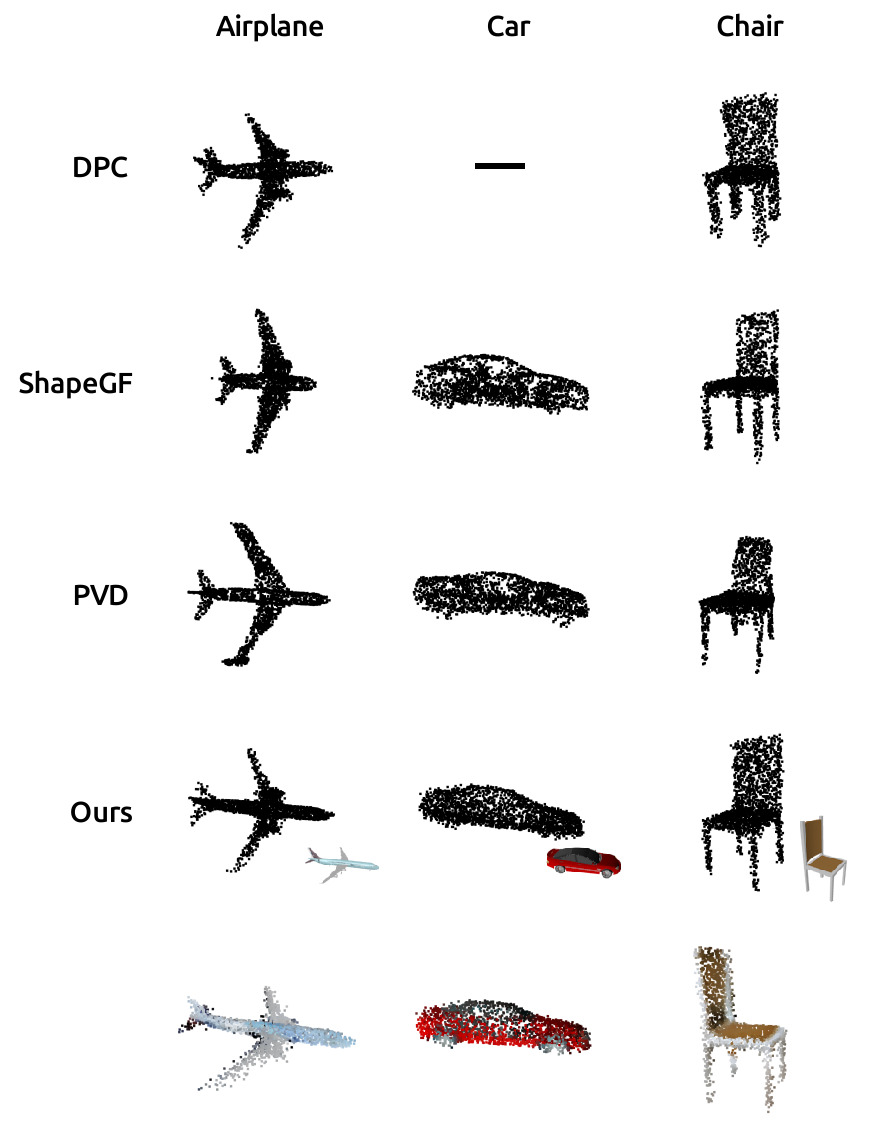}
\caption{Geometric point clouds generation results of different methods. It is noticed that DPC method does not provide pretrained model on subset of Car data. For our results, the reference image is illustrated on bottom-right, and the reconstructed colored point clouds are shown in the last row.}
\label{compare}
\end{figure}

\textbf{3D consistency of reconstruction from different views.} In this experiment, we aim to evaluate the 3D consistency of the reconstructed point clouds conditioned on different views of the same object. The experimental results are shown in Fig.~\ref{cons}. The colored point clouds are reconstructed from images from 3 distinct views. From Fig.~\ref{cons}, we can find that the proposed method can produce consistency 3D colored point cloud shape from images captured from different viewpoints. For example,  the wheels of the car can be correctly reconstructed from different views, and the airplanes reconstructed from distinct views are consistency in both geometric and color. 
However, when there is only the color of the back of the chair available as shown in the second row, the information of the seat cushion is naturally missing, and the predicted color is consistent with the back of the chair.

\textbf{Colored point clouds reconstruction from wild images.} In this experiment, we evaluate the robust of the proposed point cloud reconstruction method. Despite of the piece-wise smooth reference image provided in ShapeNet dataset, we adopt wild images in Pix3D~\cite{sun2018pix3d} dataset to validate the effectiveness of the proposed method.
The experimental results are shown in Fig.~\ref{pix}.
Although the proposed method has not seen such wild images at all during training stage, the method can also produce reasonable colored point cloud to some extent. The results in Fig.~\ref{pix} validate the robust of the proposed method effectively.

\textbf{Comparing with baselines.}
As the existing diffusion-based point clouds generation methods~\cite{cai2020learning,luo2021diffusion,zhou20213d} cannot predict color information, we compare the proposed method with the baselines on generated geometric point clouds without color. Due to that the proposed method reconstructs the point cloud conditioned on a reference image, and the baselines are completely generative methods, it is difficult to compare the performance numerically in fair case. Therefore, we compare the performance of the proposed method against baselines visually.

Fig.~\ref{compare} shows that the point cloud shapes produced by our method and the baselines. From visual inspection, the point clouds produced by our method are of high quality, with few outliers appearing, and are highly competitive. From the colored point cloud predicted by the proposed method as shown in the last row of Fig.~\ref{compare}, our method can learn more accurate 3D shape benefiting from both the geometric and  semantic color consistency. 

\textbf{Quantitative evaluation of reconstruction quality.} In this experiment, we evaluate the reconstruction quality of the point clouds quantitatively. Two metrics including Chamfer Distance (CD) and Earth Mover’s Distance (EMD) are employed in this paper.

As PVD~\cite{zhou20213d} conducted shape completion with the guidance of partial 3D shape, while DPC~\cite{luo2021diffusion} is a generative model guided by a latent shape code, and ShapeGF~\cite{cai2020learning} is a completely generative model, it is not fair to compare the reconstruction results of different methods. Therefore, we only list the image-guided reconstruction scores of the proposed method. The CD and EMD metrics are computed on the whole test dataset composed of 1501 point clouds, and the average scores are shown in Table~\ref{error}. The quantitative evaluation results show that the proposed method achieve competitive performance on 3D shape reconstruction.

\subsection{Ablation study.}
The main framework of the proposed method is composed of three modules, shape prediction module, color prediction module and NeRF-like rendering module. Among these three modules, shape prediction module and color prediction module are trained simultaneously, while NeRF-like rendering module does not have any trainable parameters. The total training objective as shown in Eq.~\ref{loss_global} is composed of two types of losses, geometric loss $\mathcal{L}_{geo} + \mathcal{L}_{cham}$ and color loss $\mathcal{L}_{rgb}$. 

\begin{figure}
\centering
\includegraphics[width=1\linewidth]{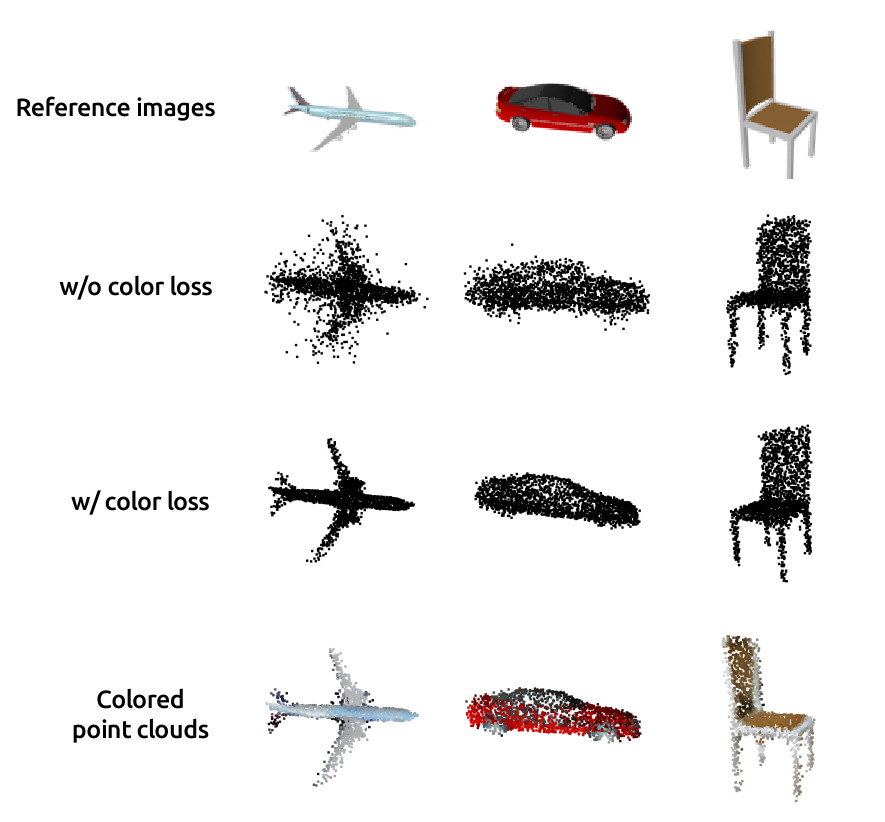}
\caption{Ablation study of reconstruction results without/with color prediction module.}
\label{ablation}
\end{figure}

Without color prediction module, one can also accomplish 3D point cloud reconstruction conditioned on the latent shape feature $\mathbf{S}$ according to our framework. In order to evaluate the role of color prediction module, we conduct an ablation experiment. The reconstruction results by methods without/with color prediction module are shown in Fig.~\ref{ablation}. From experimental results, we can find that the reconstruction results without the constraint of color similarity can produce an approximate 3D shape, however, numerous outlier points are generated as shown in the second row of Fig.~\ref{ablation}. The geometric point cloud generated with color loss is shown in the third row, and the corresponding colored shape is shown in the last row. It is easy to find that the proposed method with color prediction module improves the reconstruction quality obviously and dramatically. The experiment validates the important role of the proposed color prediction module. The average quantitative metrics are computed on the whole test data and are shown in Table~\ref{error}. It is obvious to find that the quality of reconstruction results without color constraint dropped dramatically.


\begin{table}[t]
\centering
\caption{Ablation experimental error indicators, values in parentheses are differences from the original method. CD is multiplied by $10^3$, and EMD is multiplied by $10^2$.}
\begin{tabular}{c|cc|cc}
\hline
\multirow{2}{*}{} & \multicolumn{2}{c|}{w/ color loss} & \multicolumn{2}{c}{w/o color loss} \\ \cline{2-5} 
                  & \multicolumn{1}{c|}{CD($\downarrow$)}    & EMD($\downarrow$)   & \multicolumn{1}{c|}{CD($\downarrow$)}    & EMD($\downarrow$)    \\ \hline
chair             &  \multicolumn{1}{c|}{1.333 }      &  3.960    & \multicolumn{1}{c|}{1.407}      &4.061        \\ 
airplane          & \multicolumn{1}{c|}{0.521}&2.874&\multicolumn{1}{c|}{1.273}      &4.518        \\ 
car               & \multicolumn{1}{c|}{0.917}      &   3.176    & \multicolumn{1}{c|}{1.303}      &3.799        \\ \hline
\end{tabular}
\label{error}
\end{table}

\section{Conclusion}
A novel 3d colored point cloud reconstruction method from a single RGB image is proposed in this paper. The proposed method aims to predict both the geometric and the color of the point cloud with respect to a reference RGB image through a diffusion probability model. As far as the authors know, it is the first diffusion model designed for 3D colored point cloud reconstruction. 
The proposed method is composed of three modules, shape prediction module, color prediction module and NeRF-like rendering module. 
Shape prediction module is designed to reconstruct the geometric information of point cloud, while color prediction module is to predict the color of each point. NeRF-like rendering module aims at projecting the colored point cloud predicted by the former two modules to 2D image space to guide the training. Experimental results demonstrate that the proposed method achieve competitive performance on colored point cloud reconstruction.

\bibliographystyle{abbrv-doi}

\bibliography{template}
\end{document}